\documentclass[journal=jacsat,manuscript=article]{achemso}
\usepackage{chemformula} 
\usepackage[T1]{fontenc} 
\usepackage{hyperref}

\author{Chengchun Liu}
\affiliation[Peking University]
{School of Materials Science and Engineering, Peking University, Beijing}
\author{Fanyang Mo}
\email{fmo@pku.edu.cn}
\affiliation[Peking University]
{School of Materials Science and Engineering, Peking University, Beijing}
\alsoaffiliation[Peking University Shenzhen Graduate School]
{AI for Science (AI4S)-Preferred Program, Peking University Shenzhen Graduate School, Shenzhen}

\title[An \textsf{achemso} demo]
  {Infrared Spectra Prediction for Diazo Groups Utilizing a Machine Learning Approach with Structural Attention Mechanism\footnote{A footnote for the title}}

\abbreviations{IR,NMR,UV}

\begin{document}

\begin{abstract}
  Infrared (IR) spectroscopy is a pivotal technique in chemical research for elucidating molecular structures and dynamics through vibrational and rotational transitions. However, the intricate molecular fingerprints characterized by unique vibrational and rotational patterns present substantial analytical challenges. Here, we present a machine learning approach employing a Structural Attention Mechanism tailored to enhance the prediction and interpretation of infrared spectra, particularly for diazo compounds. Our model distinguishes itself by honing in on chemical information proximal to functional groups, thereby significantly bolstering the accuracy, robustness, and interpretability of spectral predictions. This method not only demystifies the correlations between infrared spectral features and molecular structures but also offers a scalable and efficient paradigm for dissecting complex molecular interactions. 
\end{abstract}

\section{Introduction}

Infrared (IR) spectroscopy constitutes an essential tool in the domain of chemical research, facilitating profound insights into molecular architectures and functional groups via the interrogation of vibrational and rotational transitions \cite{10.1021/acs.chemrev.9b00813}. The characteristic capability of IR spectroscopy to act as a distinctive molecular fingerprint augments its applicability in the identification of chemical compounds, differentiation among analogous entities, and the elucidation of complex mixtures \cite{stuart2004infrared}. Beyond the realm of structural identification, IR spectroscopy affords the capability for the real-time monitoring of chemical reactions and conformational shifts, thereby offering an expansive perspective on molecular dynamics \cite{10.1002/chem.201500416,10.1021/ja805922b}. The formulation of theoretical models that adeptly balance accuracy with computational efficiency represents a formidable challenge, yet it is imperative for the progressive advancement of the chemical sciences.

Advancements in applying machine learning (ML) models to IR spectrum prediction have been substantial and demonstrate promising outcomes \cite{10.1021/acs.jpca.1c10417}. Schütt et al. have led the way in harnessing high-dimensional neural network potentials to expedite ab initio molecular dynamics (AIMD) simulations for IR predictions \cite{10.1021/acs.jctc.8b00908,10.1063/5.0138367}. The neural networks n2p2 and FieldSchNet estimate IR spectra by computing molecular potential energies and dipole moments \cite{10.1021/acs.jctc.8b01092}. Intriguingly, n2p2 integrates symmetry functions in its input layer, whereas FieldSchNet employs learned radial interaction functions to refine atomic representations \cite{10.1038/s41467-023-42148-y}. Wang et al. have introduced a neural network model for predicting IR spectra based on Morgan fingerprints, incorporating bulldozer distance as both a metric and loss function \cite{10.3847/1538-4357/abb5b6}. Jiang et al. developed a novel machine learning protocol for rapidly predicting the amide I infrared (IR) spectra of various proteins, using key structural descriptors, which aligns well with experimental results \cite{10.1021/jacs.0c06530}. Their method overcomes the tedious and expensive process of protein secondary structure determination from IR spectra, traditionally reliant on quantum-mechanical calculations in fluctuating environments. This cost-effective approach not only distinguishes protein secondary structures but also probes atomic structural variations with temperature and monitors protein folding, thereby modeling the relationship between protein spectra and their biological/chemical properties.

Despite their impressive predictive accuracy, these data-driven ML models often lack transparency, posing challenges in deriving chemically reasonable explanations from their results \cite{10.1021/acs.chemrev.3c00189,10.1038/s42256-020-00236-4}. Moreover, models predicated on theoretically calculated molecular descriptors are not highly scalable and necessitate considerable computational resources \cite{10.1021/acs.chemrev.0c01111}. In contrast to purely data-driven models, it is established that integrating human knowledge into data significantly enhances model interpretability and generalizability \cite{9903342}. Zhang et al. have proposed an innovative end-to-end federated learning framework \cite{10.1145/3336191.3371790}. This framework overcomes existing limitations by extracting structured knowledge from differentiable path-based recommendation models, achieved without incurring additional computational overhead. Voorhis et al. introduced an atom-based Bootstrap Embedding (BE) method, significantly enhancing its performance in capturing valence electron correlation in medium-sized molecules \cite{10.1021/acs.jpclett.9b02479}. This innovative approach promises scalable, highly accurate electron correlation solutions for large molecules.

Domain knowledge indicates that the primary factor influencing the infrared absorption value of a functional group is the information about adjacent atoms and bonds \cite{10.1021/cr500013u}. In light of this, we have developed a descriptor utilizing the Structural Attention Mechanism (SAM), which strategically prioritizes chemical information proximal to functional groups. It is crucial to highlight that SAM differs fundamentally from the self-attention mechanism \cite{schwaller2019molecular}. Models grounded in self-attention, such as the Transformer, are often categorized as having limited interpretability \cite{NIPS2017_3f5ee243}. This classification arises mainly due to their extensive parameterization and intricate internal workings, which obscure the rationale behind specific model decisions \cite{10.5555/3586589.3586709}. In contrast, our developed SAM is a feature engineering-based attention mechanism, while the self-attention mechanism falls under algorithmic attention mechanisms. The former offers a balance of robust model performance and high interpretability, whereas the latter, despite ensuring performance, often falls short in terms of interpretability.

Diazo compounds, known for their versatility in organic synthesis, serve as precursors for 1,3-dipoles, nucleophiles, and carbenes \cite{10.1142/q0261,978-0-471-13556-2}. However, their inherent instability and high reactivity render them challenging for spectroscopic analysis and direct experimentation \cite{10.1021/acs.oprd.9b00422,10.1039/C5CS00902B}. We posit that ML algorithms can surmount these challenges by deducing relationships between molecular structure and properties from valuable experimental data on IR spectra of diazo groups. In this study, we utilize the diazo group as a case study to construct an IR spectrum prediction model that emphasizes model interpretability through knowledge embedding. We demonstrate that this approach is extendable beyond the data range used in model training, facilitating generalized predictions of IR spectra for a diverse array of molecules across extensive chemical spaces. This method markedly surpasses the current theoretical and experimental limitations in spectroscopy in terms of accuracy, speed, and system size, paving the way for predicting vibrational and rotational spectra and comprehending complex intra- and intermolecular interactions. Remarkably, the model aligns with chemical intuition in its decision-making process.

\section{Results and discussion}

\subsection{Dataset construction and feature engineering}

In this research, we compiled a comprehensive dataset encompassing 1,827 diazo compounds, characterized by their molecular structures through Simplified Molecular Input Line Entry System (SMILES) notations and their infrared absorption spectra obtained from experimental measurements. It is essential to emphasize that our methodology was firmly grounded in experimental observation, explicitly avoiding the use of theoretical models in favor of direct empirical evidence. This comprehensive dataset was extracted from 297 scholarly articles utilizing a regular expression-based method for intelligent text retrieval, with the detailed methodology outlined in the Supplementary Information (SI). The collected data notably features the characteristic absorption peak frequencies of diazo groups in the infrared spectrum, spanning a range from a minimum of 2000 to a maximum of 2200 (Figure 1A).

\begin{figure}
\centering
  \includegraphics[width=\textwidth]{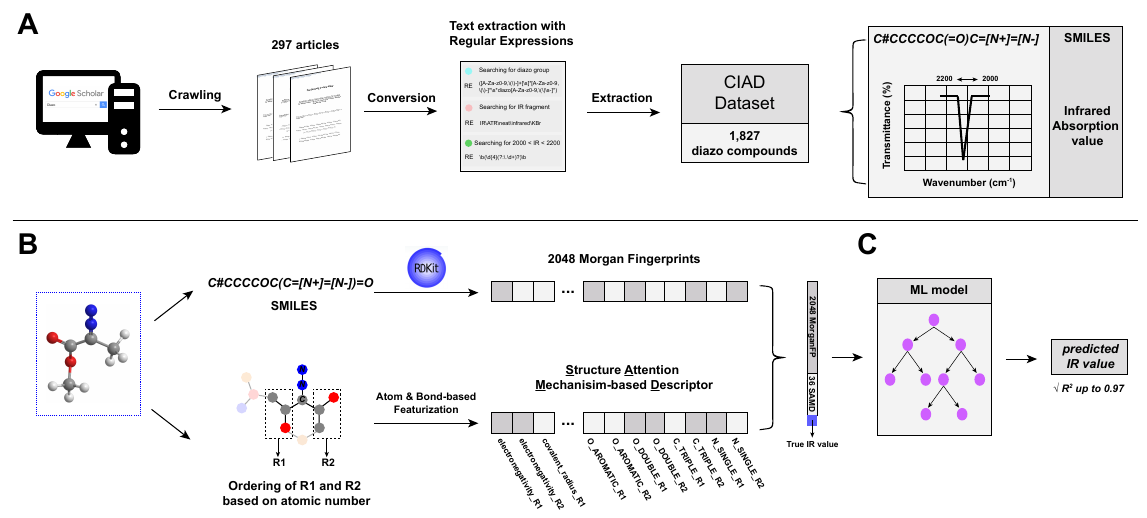}
  \vspace{-1em}
  \caption{\textbf{The workflow of this study.} \textbf{(A)} Database creation workflow. \textbf{(B)} Feature engineering workflow. \textbf{(C)} Machine learning prediction.}
\end{figure}

To ensure a broad representation of data, the dataset encompasses diazo groups in conjunction with a variety of other chemical groups.  For analytical purposes, two types of descriptors were employed: the Structure Attention Mechanism-based Descriptor (SAMD) and the 2048-bit Morgan fingerprint (MorganFP), derived from the SMILES sequences of the molecules (Figure 1B).

The SAMD is bifurcated into two components: first, it assesses the electronegativity and covalent radius of the atom directly bonded to the diazo group; second, it considers the atoms (including non-metallic and metalloid elements such as I, Br, Cl, S, P, Si, F, O, C, N, B, and H) and various bond types (single, double, triple, and aromatic bonds) linked to this first atom. After eliminating combinations deemed implausible under valence bond theory, a total of 36 viable combinations were identified and numerically encoded for analysis. The finalized dataset comprised 2,084 descriptors, which underwent normalization before their application in model development. In this study, the CIAD dataset is partitioned into an 80\% training set and a 20\% test set using a random allocation method. The predictive performance of each model is quantitatively assessed through a rigorous process of cross-validation and subsequent testing phases. To evaluate the accuracy and efficacy of the derived models, two key performance metrics are employed: the coefficient of determination (R$^2$) and the root mean square error (RMSE). These metrics provide a comprehensive measure of the models' predictive accuracy and their capability to generalize to unseen data.

\subsection{Model construction and performance evaluation}

In this study, we conducted a comprehensive assessment of several leading machine learning algorithms, encompassing Random Forest (RF), LightGBM (LGB), XGBoost (XGB), Bayesian Ridge Regression (Bayesian), Gradient Boosting (GB), and CatBoost (CB). Our comparative analysis was centered on evaluating the algorithms' efficacy using R$^2$ under cross-validation, juxtaposed against RMSE metrics.

Our results revealed that decision tree-based algorithms, particularly RF, LGB, XGB, and CB, exhibited superior predictive abilities, as indicated by their R$^2$ scores exceeding 0.95 (Figure 2A). We observed a notable diminution in performance when relying exclusively on MorganFP, which relies on 2-dimensional structural information. Detailed performance metrics for these algorithms are provided in Figure S4. These findings underscore the structure attention mechanism's (SAM) effectiveness in feature engineering, especially in capturing essential chemical data pivotal for determining the spectral absorption frequencies of molecular groups.

To augment the predictive precision of our model, we adopted an ensemble learning approach, culminating in the creation of a mixture model. This model amalgamates stacking and voting methods from ensemble learning, employing multiple regression models (RF, GB, CB, LGB, and XGB) as base models and Bayesian Ridge Regression as the meta-model. Moreover, we integrated the predictions from the stacked regressor with the highest-performing base model, XGB, using a voting regressor (Figure S5). This integration aimed to enhance overall prediction accuracy and bolster the model's robustness. The mixture model demonstrated remarkable accuracy, with its cross-validation R$^2$ score reaching 0.969 (Figure 2B). This significant performance improvement, compared to individual models, underscores the complexity and non-linear nature of the descriptors utilized in this study for predicting characteristic group peaks in infrared spectra.

Further exploration into the interplay between model predictive capacity and input data involved using varying quantities of compounds for training while maintaining a constant test set. The learning curve of the mixture model indicated a correlation between increased training data volume and improved performance (Figure 2C). Notably, with a data volume of 25\% (456 diazo compounds), the R$^2$ score attained 0.90. The data volume reached a plateau at 65\% (1187 diazo compounds), aligning with the cross-validation curve, signifying comprehensive training and robust generalization capability. This suggests a significant physical correlation between the compounds' structures and properties and their infrared absorption characteristics. The proposed mixture model, leveraging a structure attention mechanism (SAM), demonstrates its capacity to learn effectively with as few as 456 compounds, thereby predicting infrared characteristic absorption values with considerable accuracy.

\begin{figure}
\centering
  \includegraphics[width=\textwidth]{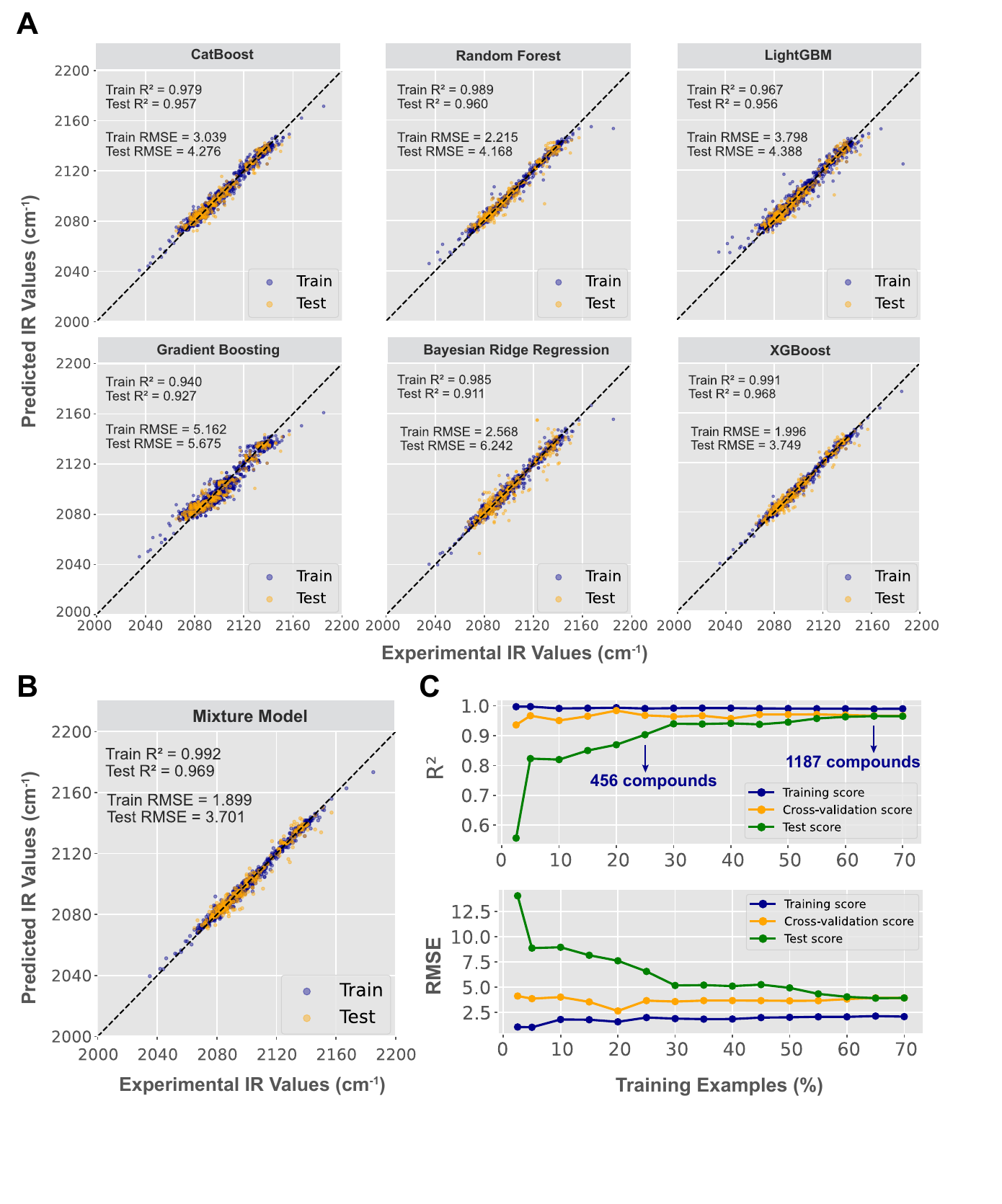}
  \vspace{-1em}
  \caption{\textbf{Model construction and performance evaluation.} \textbf{(A)} Evaluation of model performance based on some common machine learning algorithms. \textbf{(B)} Performance evaluation of mixture model. \textbf{(C)} Evaluation of the impact of different training data on model performance.}
\end{figure}

\subsection{Model robustness and generalization analysis}

In the realm of predictive modeling, it is imperative to ascertain the robustness of a model under varying conditions. Previous studies, including our own, have intuitively established that the efficacy of machine learning models frequently hinges on the similarity between the structures used in training and those in prediction, particularly regarding molecular structures \cite{10.1038/s41467-023-38853-3}. This investigation focuses on the use of ML models for predicting the characteristic infrared absorption peaks of diazo groups, alongside the chemical structure information of adjacent groups. Notably, this approach enables the prediction of hitherto unreported infrared spectral data. To this end, we have explored the generalization capabilities of this model across diazo compounds exhibiting varying degrees of structural similarity.

The study employs the Tanimoto similarity coefficient as a metric to quantify the similarity between two-dimensional molecular structures, with values ranging from 0\% to 100\%. For each model in cross-validation, we calculate the Tanimoto coefficient between each test set diazo compound and all diazo compounds in the training set. We have established multiple similarity thresholds (90\%, 85\%, 80\%, 75\%, 70\%, 65\%, 60\%, and 50\%) to delineate different levels of similarity. Consequently, the data were segmented into eight groups, with each group containing test set molecules that shared at least one similar molecule in the training set above a specified similarity threshold. The group sizes were as follows: n90 = 178 (48.5\%), n85 = 186 (50.7\%),  n80 = 206 (56.1\%), n75 = 241 (65.7\%), n70 = 277 (75.48\%), n65 = 301 (82.0\%), n60 = 318 (86.7\%), and n50 = 357 (97.3\%). Notably, 13.3\% of the diazo compounds exhibited no more than 60\% similarity to any molecule in the training set. Figure 3A elucidates the predicted R$^2$, RMSE, and the mean absolute error (MAE) for these groups, affirming a strong correlation between molecular similarity and generalization capability. As similarity diminishes, the predictive accuracy of the model markedly decreases. For molecules with over 90\% similarity, the model demonstrated commendable performance, exhibiting an R$^2$ of 0.981 and an RMSE of 2.661. Conversely, for molecules with over 50\% similarity, these values were 0.970 (R$^2$) and 3.691 (RMSE), respectively. Intriguingly, the R$^2$ for molecules with less than 50\% similarity still reached 0.897, as shown in Figure S6.

Moreover, model performance is often contingent upon data quality, with noise being a pivotal factor. Noise may stem from measurement errors, data processing, or external environmental uncertainties. To elucidate the predictive prowess of the proposed mixture model, we investigated the effects of data volume and noise. Model performance was assessed by introducing various levels of random noise into the input data, ranging from 0\% (no noise) to 100\% (extreme noise), incremented in 10\% stages. For each noise level, the model's predicted R$^2$ value was computed, and the experiment was repeated for result reliability. As depicted in Figure 3B, in the absence of noise, the model exhibited exceptional predictive capabilities (R$^2$ = 0.968). With increasing noise levels, a gradual decline in R$^2$ was observed. Notably, the model maintained an R$^2$ above 0.9 for noise levels below 50\%, demonstrating robustness. However, beyond 50\% noise, performance degradation accelerated, yet the model still maintained an R$^2$ of 0.905 at 60\% noise. This also underscores the inherent experimental errors in observational data, further corroborating the mixture model's efficacy in noise handling.

The CIAD data set contains a range of stable diazo compounds. Within the extensive chemical landscape, a significant fraction of cost-effective diazo compounds exhibit instability, presenting obstacles for direct experimental evaluation and molecular property elucidation \cite{10.1021/acs.chemrev.1c00991,10.1021/acs.accounts.0c00199}. To assess the model's ability to extrapolate, we predicted the infrared spectrum of diazomethane, a very unstable but versatile diazo compound. This assessment required a tripartite analysis, combining theoretical calculations and empirical measurements juxtaposed with model-based predictions. Theoretical analysis yielded a spectral value of 2116.2 cm$^{-1}$, while a precision-guided experimental approach determined a value of 2091.0 cm$^{-1}$ (refer to Supplementary Information for detailed methodology). The model was developed in this paper Developed under a rigorous analytical framework, the estimated spectral value is 2087.3 cm$^{-1}$. The proximity of the model's estimations to values derived through empirical methodologies underscores its predictive accuracy and its capacity to mitigate the challenges associated with the direct analysis of unstable compounds. The concordance observed between the predicted and experimental values underscores that the learning model, engineered through the incorporation of a structural attention mechanism, is adept at discerning the interrelations between molecular structure and properties, thereby exhibiting pronounced robustness and generalizability.

\begin{figure}
\centering
  \includegraphics[width=\textwidth]{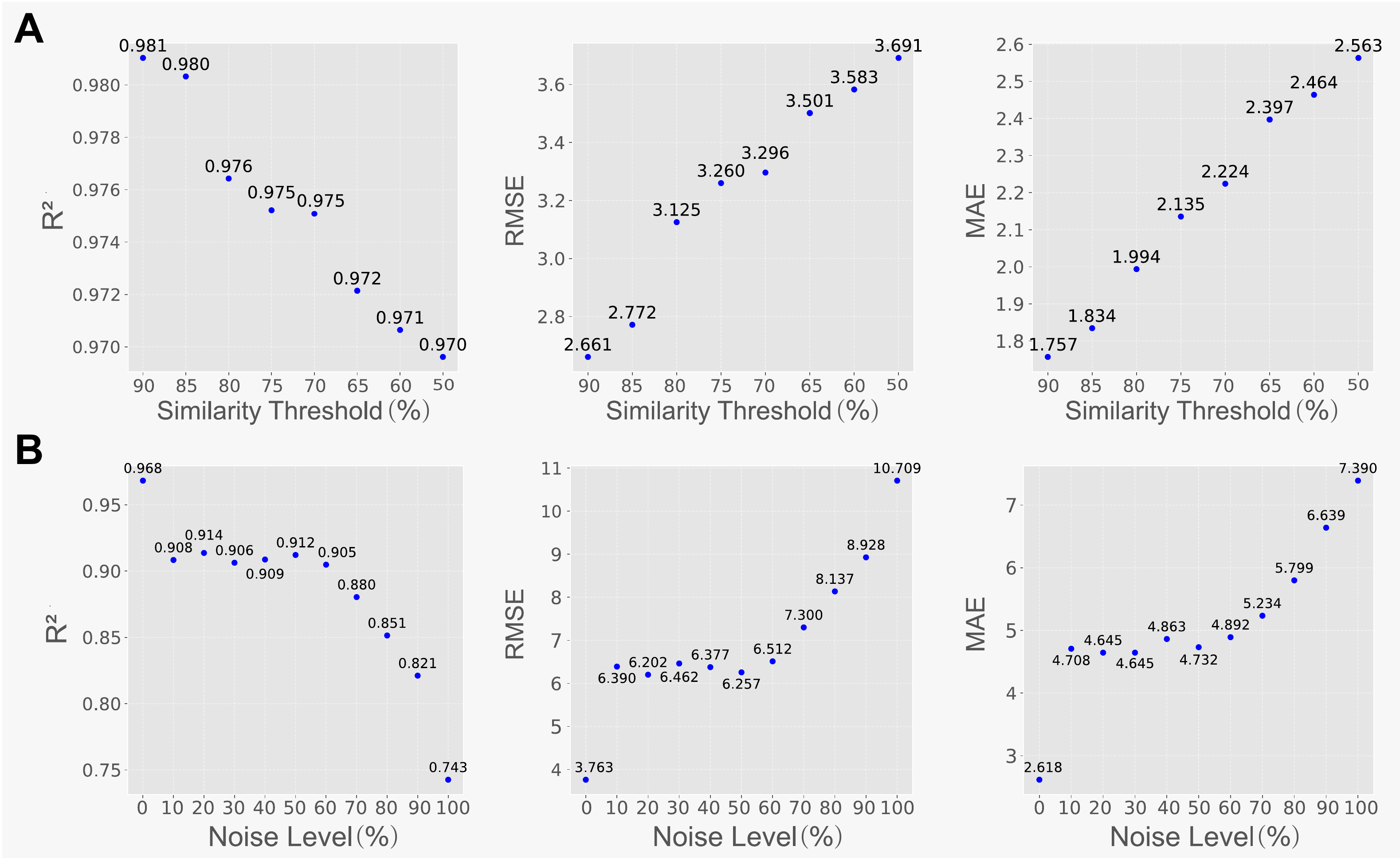}
  \vspace{-1em}
  \caption{\textbf{Model Robustness Analysis.} \textbf{(A)} The influence of similarity on the prediction model. \textbf{(B)} The influence of noise data on the prediction model.}
\end{figure}

\subsection{Model decision-making process and interpretability analysis}

The execution of model interpretability analysis is imperative for elucidating the underlying logic in model predictions. This analysis typically bifurcates into two streams: ex-ante analysis and post-event analysis. In the quest for precision and minimal error in the infrared spectrum model, the SHapley Additive exPlanations (SHAP) method is employed for interpretability \cite{lundberg2017unified}. As a post hoc interpretative tool, SHAP uniquely quantifies the contribution of each feature to the model's output, thereby facilitating a detailed and layered understanding of the ``black box'' model at both macroscopic and microscopic levels.

Feature importance analysis forms a critical juncture bridging data science theory with its practical applications, playing an instrumental role in the development of efficient and interpretable machine learning models. In this study, the feature importance was ascertained based on SHAP values, with the carbonyl descriptor emerging as the most influential, evidenced by its high average SHAP value of 6.8. This contrasts markedly with other descriptors, all of which recorded values below 1.0 (Figure 4A). A notable observation was the correlation between the number of carbonyl groups and the wave number; an increase in carbonyl groups (tending to red) corresponded to an elevated wave number, and vice versa (Figure 4B). Additionally, the atom attached to the diazo group was found to exert significant influence, with its electronegativity being directly proportional to the wave number. This relationship is further corroborated by a Pearson correlation coefficient of 0.71 between the wave number and carbonyl groups, indicating a strong positive correlation (Figure 4C).

Further examination of specific compounds, such as methyl (\textit{E})-2-diazo-5-(4-methoxyphe\-nyl)pent-4-enoate, revealed nuanced feature engineering based on the structural attention mechanism. This involved segmentation into two chemical space domains, R1 and R2, based on atomic number size, followed by quantitative characterization according to atom types and chemical bonds (Figure 4D).

In the context of model decision-making, the central gray vertical line in the decision diagram represents the model's baseline value, with colored lines depicting predictions and illustrating the influence of each feature in shifting the output value above or below the average prediction. Eigenvalues adjacent to the prediction line provide a reference. The cumulative impact of SHAP values from the base value to the final predicted wave number is depicted at the graph's apex (Figure 4E), highlighting the top 10 most impactful features. The absence of a double-bonded oxygen fragment in the R2 domain (O\_DOUBLE\_R2 = 0) significantly influences model decision-making, decreasing the wave number from approximately 2092.1 to 2082.0. This effect is attributed to the electron-attracting capability of the carbon-oxygen double-bonded fragment impacting the diazo group. Conversely, the presence of a double-bonded oxygen fragment in the R1 domain (O\_DOUBLE\_R1 = 1) slightly elevates the wave number from the baseline. Moreover, the presence of two single-bond hydrogen fragments (H\_SINGLE\_R2 = 2) decreases the wave number from around 2096.6 to 2092.1, linked to the moderate electron-donating effect of the carbon-hydrogen single-bond fragment on the diazo group. The absence of a triple-bonded nitrogen fragment (N\_TRIPLE\_R2 = 0) reduces the wave number from 2098.4 to 2096.6, attributed to the electron-withdrawing effect of the carbon-nitrogen triple-bonded fragment on the diazo group. Additionally, the covalent radius of the atom linked to the R2 domain and the diazo group (colvalent\_radius\_R2 = 77) plays a role; generally, a larger covalent radius indicates a greater electron-donating propensity, and vice versa. Given the diverse heteroatoms in the dataset (e.g., H = 37, B = 82, N = 75, O = 73, Si = 111, P = 106, S = 102), a covalent radius of 77 (representing carbon atoms) necessitates a reduction in wave number from around 2101.0 to 2098.4. Additionally, the presence of a single bonded oxygen in the R1 domain (O\_SINGLE\_R1 = 1) increases the wave number, whereas its absence in the R2 domain (O\_SINGLE\_R2 = 0) decreases it. This effect is attributed to the electron-withdrawing impact of the carbon-oxygen single bond on the diazo group. 

The influence of these features on model decisions is quantifiable through force plots (Figure 4F), with the final predicted output value being 2082.0. This alignment with established chemical principles underscores the model's reliance on chemical logic and rules. In summation, constructing models with heightened chemical interpretability is not only essential for validating their predictions but also acts as a conduit between machine learning and conventional chemistry. The integration of model insights with established chemical theories not only bolsters the reliability of computational predictions but also deepens the understanding of chemical concepts. This synergy of advanced data-driven methodologies with traditional scientific knowledge heralds a transformative era in chemical research and theoretical exploration.

\begin{figure}
\centering
  \includegraphics[width=\textwidth]{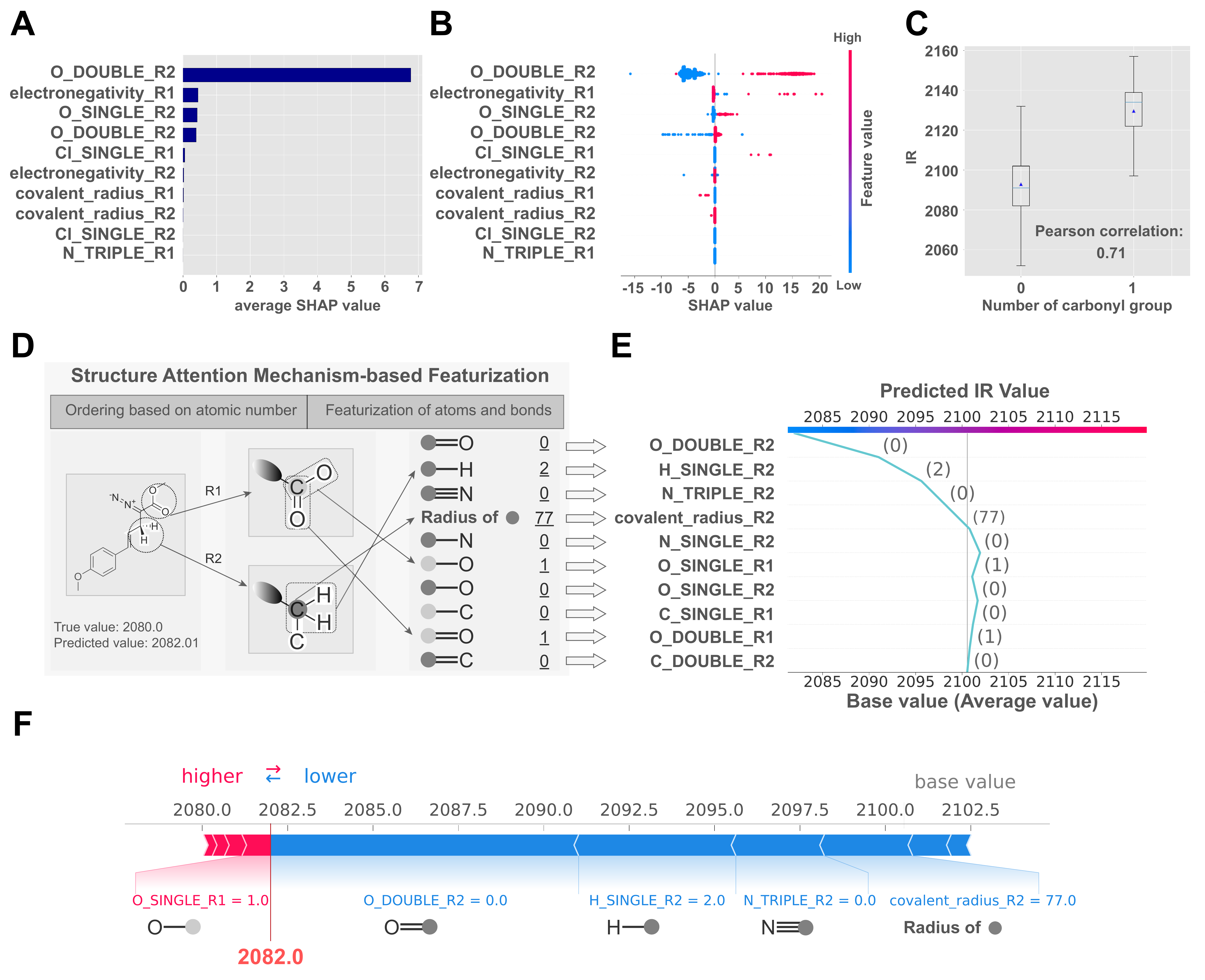}
  \vspace{-1em}
  \caption{\textbf{Model Interpretability Analysis.} \textbf{(A)} Feature importance analysis (top 10 features) based on average SHAP value. \textbf{(B)} Analysis of Sample Distribution Utilizing SHAP Values. Empirical analysis indicates that the feature O\_DOUBLE\_R2 holds predominant significance, on average, in the context of infrared absorption. Notably, its characteristic value is identified as 0 (denoted in blue), suggesting that the corresponding wavenumber is less likely to exceed the average threshold. \textbf{(C)} Pearson correlation analysis between the infrared wave number of the diazo group and the number of carbonyl groups. \textbf{(D)} Featurization of arbitrarily selected molecules utilizing a structural attention mechanism (SAM). \textbf{(E)} Model of decision-making process informed by SHAP values. This diagram presents a hierarchical analysis, beginning from the base with the mean wave number. Progressing upwards, it delineates how various features sequentially influence the model's decisions. The apex of the structure illustrates the ultimate predicted infrared (IR) value derived from this cumulative assessment. \textbf{(F)} A force diagram using SHAP values measures how different features affect the model's decisions. Features that increase the wave number are on the left, and those that decrease it are on the right. The model's final prediction is 2082.0.}
\end{figure}

\section{Conclusions}

In summary, this study demonstrates the efficacy of a machine learning approach with a structural attention mechanism(SAM) for predicting the infrared spectra of diazo groups. This method addresses existing limitations in spectral analysis, offering improved accuracy and efficiency. The integration of the structural attention mechanism is particularly notable for its enhancement of prediction capabilities, underscoring the potential of machine learning in spectroscopic applications. These findings not only contribute significantly to the field of computational chemistry but also open avenues for future research, including the model's adaptation for other complex molecular structures. The broader implications of this research extend to advancing organic synthesis and deepening our understanding of molecular interactions in various chemical contexts.

\section{ASSOCIATED CONTENT}

\subsection{Data Availability}

The computational models and data sets reported in this work will be made available on GitHub.

\subsection{Supporting Information}

The Supporting Information is available free of charge at XXX.

\section{AUTHOR INFORMATION}

\subsection{Corresponding Author}

\textbf{Fanyang Mo -- }
\textit{School of Materials Science and Engineering, Peking University, Beijing 100871, China; AI for Science (AI4S)-Preferred Program, Peking University Shenzhen Graduate School, Shenzhen 518055, China}; 
ORCID: 0000-0002-4140-3020; 
Email: \href{mailto:fmo@pku.edu.cn}{fmo@pku.edu.cn}

\subsection{Author}
\textbf{Chengchun Liu -- }
\textit{School of Materials Science and Engineering, Peking University, Beijing 100871, China}

\subsection{Author Contributions}

F.M. and C.L. conceived the project, designed the method, and analyzed the results. C.L. organized and prepared structural data. F.M. supervised the project. F.M. and C.L. wrote the paper.

\subsection{Notes}

The authors declare no competing financial interest.

\section{ACKNOWLEDGMENTS}

This work was financially supported by the National Natural Science Foundation of China (Grant Nos. 22071004, 21933001 and 22150013). We thank the High-Performance Computing Platform of Peking University for machine learning model training.

\bibliography{achemso-demo}

\section{TOC Graphic}

\begin{figure}
\centering
  \includegraphics[width=\textwidth]{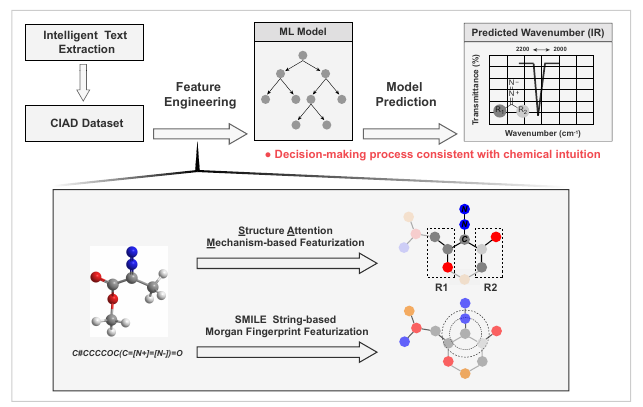}
  \vspace{-1em}
\end{figure}

\end{document}